%% file: main_arxiv.tex
\documentclass{article}
\input{preamble}

\begin{document}
\input{abstract}
\input{introduction}
\input{problem}
\input{algorithm}
\input{experiments}
\input{related_work}

\input{conclusion}
\input{acknowledgements}

\bibliography{ref}
\bibliographystyle{icml2021}

\input{supplement_abstract}
\input{supplement_appendix}

\end{document}

%% file: preamble.tex
\usepackage{microtype}
\usepackage{booktabs}
\usepackage{hyperref}
\usepackage{color}

\usepackage{graphicx}
\usepackage{caption}
\usepackage{subcaption}

\usepackage{amsmath,amsfonts,amssymb,amsthm}
\usepackage{float}
\usepackage[mathscr]{euscript}
\usepackage[dvipsnames]{xcolor}
\usepackage{natbib}
\usepackage[ruled]{algorithm}
\usepackage[noend]{algorithmic}
\usepackage{makecell}
\usepackage{tikz}
\usetikzlibrary{fit,calc}

\usepackage[accepted]{icml_arxiv_2021}

\include{commands}

\icmltitlerunning{Adapting the Function Approximation Architecture in Online Reinforcement Learning}

%% file: commands.tex
\newcommand{\fbf}{\mathbf{f}}

\newcommand{\obf}{\mathbf{o}}

\newcommand{\wbf}{\mathbf{w}}
\newcommand{\xbf}{\mathbf{x}}
\newcommand{\ybf}{\mathbf{y}}
\newcommand{\zbf}{\mathbf{z}}

\newcommand{\Abf}{\mathbf{A}}

\newcommand{\Mbf}{\mathbf{M}}

\newcommand{\Ccal}{\mathcal{C}}

\newcommand{\Ncal}{\mathcal{N}}
\newcommand{\Ocal}{\mathcal{O}}
\newcommand{\Pcal}{\mathcal{P}}

\newcommand{\Rcal}{\mathcal{R}}
\newcommand{\Scal}{\mathcal{S}}

\newcommand{\Ucal}{\mathcal{U}}

\newcommand{\ellbf}{\boldsymbol{\ell}}

\newcommand*{\tikzmk}[1]{\tikz[remember picture,overlay,] \node (#1) {};\ignorespaces}

\newcommand{\outlineboxit}[1]{\tikz[remember picture,overlay]{\node[xshift=-.75\columnwidth, yshift=-.92\baselineskip, draw=#1, rounded corners=.2cm,fit={(A)($(B)+(.85\columnwidth,1.9\baselineskip)$)}] {};}\ignorespaces}

\colorlet{highlight}{white!30!yellow!70!}
\colorlet{comments}{white!20!black!80!}

%% file: abstract.tex
\twocolumn[
\icmltitle{Adapting the Function Approximation Architecture\\ in Online Reinforcement Learning}
\icmlsetsymbol{equal}{*}

\begin{icmlauthorlist}
\icmlauthor{John D. Martin}{equal,uofa,dm,intern}
\icmlauthor{Joseph Modayil}{equal,dm}
\end{icmlauthorlist}

\icmlaffiliation{uofa}{Department of Computing Science, University of Alberta, Canada}
\icmlaffiliation{dm}{DeepMind, Edmonton, Canada}
\icmlaffiliation{intern}{Work done during an internship}

\icmlcorrespondingauthor{John D. Martin}{jmartin8@ualberta.ca}
\icmlkeywords{Reinforcement Learning, Function Approximation, Online, Value Function, Sparse}
 \vskip 0.3in
 ]
\printAffiliationsAndNotice{\icmlEqualContribution}

\begin{abstract}
The performance of a reinforcement learning (RL) system depends on the computational architecture used to approximate a value function.
Deep learning methods provide both optimization techniques and architectures for approximating nonlinear functions from noisy, high-dimensional observations.
However, prevailing optimization techniques are not designed for strictly-incremental online updates. 
Nor are standard architectures designed for observations with an {\em a priori} unknown structure: for example, light sensors randomly dispersed in space.
This paper proposes an online RL prediction algorithm with an adaptive architecture that efficiently finds useful nonlinear features.
The algorithm is evaluated in a spatial domain with high-dimensional, stochastic observations.
The algorithm outperforms non-adaptive baseline architectures and approaches the performance of an architecture given side-channel information.
These results are a step towards scalable RL algorithms for more general problems, where the observation structure is not available.
\end{abstract}

%% file: introduction.tex
\section{Introduction}
Reinforcement learning (RL) systems have scaled to challenging problem settings by adopting function approximation architectures from deep learning~\cite{mnih2015human,silver2016mastering}.
The resulting deep RL systems often use less expert knowledge to represent a value function than their alternatives, which have historically used hand-designed, nonlinear features.
A deep RL system expresses its architecture as a large graph of differentiable computations, then optimizes its internal parameters with stochastic gradient descent and mini-batches of stored experience.
To remain tractable, large systems typically impose sparse connections from knowledge of the observation structure; for example, spatial neighborhoods guide connectivity in convolutional architectures.
Despite their benefits, it is unclear how to best deploy a deep RL system when the observation structure is unknown.
It is also unclear how to deploy deep RL when learning must occur \textit{incrementally online}, without storing transitions in a replay buffer.
Our paper studies RL systems that support both constraints.

In the strictly-incremental online RL regime considered here, a learning system is updated one transition at a time, from a single stream of experience~\cite{sutton1988learning}.
The system is not permitted to store multiple transitions, and it must cope with any temporal dependencies between consecutive transitions. 
In contrast, most deep RL systems sample transitions from a replay buffer to reduce temporal dependencies~\cite{riedmiller2005neural}, or they learn from parallel streams of experience~\cite{mnih2016async}.
Efforts to develop online supervised deep learning systems are ongoing but remain challenging~\cite{sahoo2018online}.   Separate efforts are developing backpropagation-free neural architectures~\cite{veness2020gated}. 
Instead of developing deep RL algorithms that learn online, our paper studies only how to adapt the connectivity of a nonlinear architecture. 

Typical architectures fix the connectivity when observation structure is known. 
These include convolutions~\cite{lecun1998gradient}, transformers \cite{vaswani2017attention}, and graph neural networks~\cite{scarselli2008graph}.
However, there are situations when learning from observations with unknown structure will be required (e.g. fusing data from many uncalibrated sensors), and one still wishes to impose a sparse structure for computational efficiency.
Prior work on learning the architecture's connectivity has examined small observation spaces~\cite{fritzke1995}, searching a combinatorially-large space of graphs~\cite{zoph2017neural, stanley2019designing, mahmood2013representation, rahman2021towards}, and mini-batch methods that learn to sparsify a dense architecture~\cite{neyshabur2020towards,gale2019state}. 

We are motivated by the increasing ability of machine learning algorithms to reduce reliance on expert knowledge~\cite{halevy2009unreasonable}.
Furthermore, we would like to take a step towards sparse RL architectures that can learn online, from high-dimensional, stochastic observations of unknown structure.
A key idea underlying our approach is that RL systems can learn useful sparse structure from a collection of parallel auxiliary predictions.  

This paper proposes an RL algorithm that adapts an architecture's connectivity with information from learned auxiliary predictions.
In principle, the algorithm can be applied to many architectures, but our study focuses on one that is well-suited to the online setting.
The proposed algorithm specifies auxiliary prediction objectives as General Value Functions (GVFs).
Each GVF selects a set of informative observation components, which we call a \textit{neighborhood}.
The components in each set are then nonlinearly combined to form useful features for a main value function.
The algorithm is validated in a challenging synthetic domain with high observation noise.  
Our results show the algorithm can learn to adapt the approximation architecture without incurring substantial performance loss, while also remaining computationally tractable. 
We highlight the following contributions. 

\textbf{Online adaptive architecture:} 
Using many parallel auxiliary learning objectives, our architecture dynamically connects observations to a set of filter banks to form useful nonlinear features.  
 
\textbf{Useful neighborhoods:} The proposed algorithm is shown to compute sparse neighborhoods that perform comparably well to neighborhoods formed from side-channel distance information, and it substantially outperforms static baseline architectures. 

\textbf{Reduced architectural bias with conventional data and computational resources:} 
In a domain of noisy observations with thousands of dimensions, useful neighborhoods are found within five-million time steps of experience, after running on a single GPU for two hours.

%% file: problem.tex
\section{Problem Setting}
The environment is defined as a non-episodic discounted Markov reward process $(\mathcal{S},p, p_1, \gamma)$ with states $s\in \mathcal{S}$, transition distribution $p(S',R|S=s)$, initial state distribution $p_1(S_1)$, and a discount factor $\gamma\in[0,1)$.
It produces a trajectory of alternating states and rewards: $S_1, R_2,  S_2, R_3, S_3 \ldots$. 
From this process, the return at time $t\in \mathbb{N}$ is defined as the discounted sum of future rewards,
\begin{equation}
  \label{eq:return}
G_t \equiv R_{t+1} + \gamma R_{t+2} + \gamma^2 R_{t+3}+ \ldots.
\end{equation}
The \textit{value function} \citep{sutton2018reinforcement} gives the expected return from a state: $v(s) \equiv \mathbb{E}[G_t | S_t = s ]$. 
 
Instead of experiencing states directly, the learner receives a stream of observation vectors and rewards.
Here, an observation vector is generated at each time step $t$ by an unknown random function of the state: $\mathbf{o}_t \equiv o(S_t) \in \mathbb{R}^d$. 
The learner's only knowledge of the observation function and of the environmental dynamics comes from this single stream of experience.
With no direct access to the environment's state, the learner forms an approximate value function to estimate the expected return. 
In particular, the approximation is defined as a function of a feature vector $\xbf_t \in \mathbb{R}^\ell$,
 \begin{align}
  \hat{v}( \xbf_t ; \wbf_t) &\equiv \wbf^\top_t \xbf_t, &  \hat{v}( \xbf_t; \wbf_t) &\approx v(S_t).
  \label{eq:value}
 \end{align}
In the online setting considered in this work, the learner can incrementally update the weights $\mathbf{w}_t$ with a temporal difference learning algorithm such as $\text{TD}(\lambda)$.
 
The feature vector is computed as a function of the observation, which can be linear or nonlinear.
Together with a weight update rule for \eqref{eq:value}, this function that computes features determines the prediction system's approximation architecture.

\section{An Approximation Architecture}
Our work is concerned with problems where linear approximations are inadequate, and a nonlinear approximation architecture is required.
In particular, we study cases where the observation dimension is moderate (e.g. $d > 1000$), each individual observation is simple but noisy (e.g. $\obf_t \in \{0,1\}^d$), and a descriptive feature vector may have many components ($\ell > d$).

\begin{figure}
	\centering
	\includegraphics[width=\columnwidth]{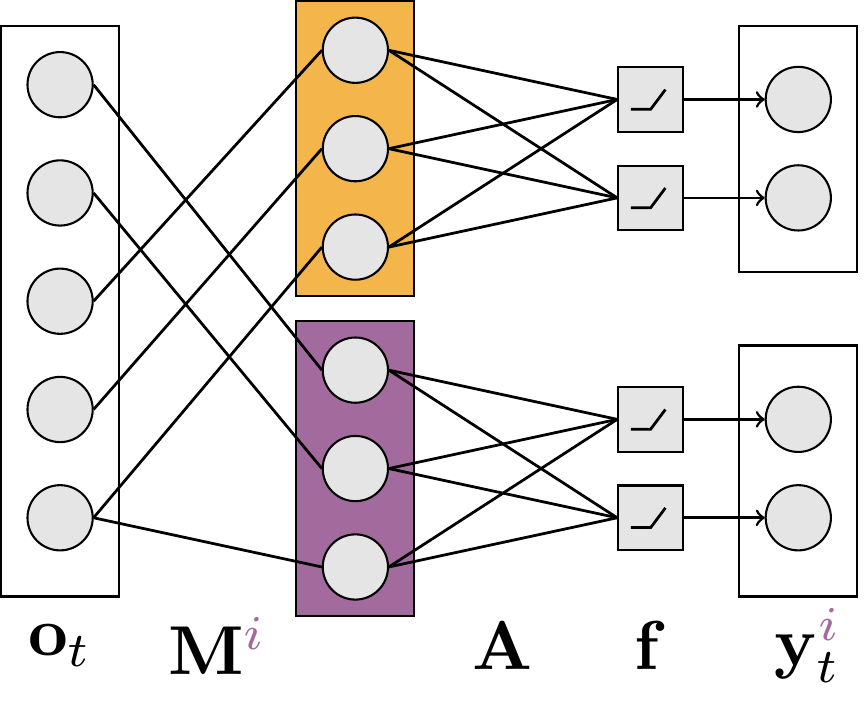}
	\caption{\textbf{Nonlinear features:} Two neighborhoods comprised of three observation components are transformed with $\Abf$ then passed through a nonlinearity $\fbf$ to generate two sets of two nonlinear features. The two neighborhoods (purple and orange) have different selection matrices and generate different nonlinear features.}
	\label{fig:feature_bank}
\end{figure}

We consider an architecture that computes nonlinear features from a wide, shallow (one-layer) network (Figure \ref{fig:feature_bank}).
It forms sparse connections from the input observations by selecting subsets we call \textit{neighborhoods}.  
Next, the architecture applies a set of linear filters with fixed parameters, followed by a nonlinear transformation.
This is similar to a convolutional neural network layer \citep{lecun1998gradient}, which imposes sparse connections with prior knowledge of the observation's spatial structure. 
A typical convolutional layer fixes its neighborhoods (i.e. the inputs applied to the kernel), while its filter parameters are optimized with stochastic gradient descent.
Here neighborhoods are treated as an architectural element for sparsity, and they are obtained by masking out a subset of the observation vector.
In the next section we describe how to compute these masks without \textit{a priori} knowledge of the observation structure.   

The nonlinear feature vector from the $i$-th neighborhood is computed as a composition of three functions,
\begin{align}
	\ybf_t^{i} \equiv  \fbf( \Abf \Mbf^{i} \obf_t).
	\label{eq:feature_bank}
\end{align}
First is a neighborhood selection matrix $\Mbf^{i}: \mathbb{R}^d \rightarrow \mathbb{R}^k$, then a linear projection $\Abf: \mathbb{R}^k \rightarrow \mathbb{R}^n$, and finally a nonlinearity $\fbf \colon \mathbb{R}^n\rightarrow \mathbb{R}^n$. 
The neighborhood selection matrix $\Mbf^{i}$ is an orthogonal rank-$k$ matrix of zeros and ones that provides an ordered selection of $k$ elements of the observation (so $k \leq d$). 
The linear projection $\mathbf{A}$ can be thought of as a set of filters, which can also include a bias unit.
The function $\fbf$ applies a fixed nonlinearity $f\colon \mathbb{R}\rightarrow \mathbb{R}$ to each element of its $n$-dimensional input: $\fbf(\zbf) = (f(\zbf_1), ... , f(\zbf_n))$.  

The full feature vector, $\xbf_t$, contains nonlinear features from $m$ neighborhoods, $\Mbf^i\obf_t$, and the current observation,
\begin{align}
 \xbf_t \equiv \text{concatenate}(\obf_t, \ybf_t^1,\ldots,\ybf_t^m).
 \label{eq:stateupdate}
\end{align}

Although this architecture is simple, it remains expressive. 
It can represent a dense architecture with $\Mbf={\mathbf I}^{d\times d}$, that selects all observation components.
It can represent a sparse architecture with $m$ neighborhoods, each with $k$ observation components.
By construction, it generalizes the shared weights and sparse connectivity of a convolutional architecture without the restriction of inputs arriving on a rectilinear grid; each $\mathbf{M}^{i}$ selects one fixed block of the image, $\mathbf{A}$ corresponds to the shared convolutional weights, and $f$ to a common nonlinearity, such as a ReLU function.
Additionally, random projections studied by \citet{sutton1993online} can be represented by choosing $n=1$ with a threshold-based nonlinearity and a different choice of $\mathbf{A}$ for each neighborhood\footnote{In this paper, we use a common $\Abf$, as we are not studying the optimization of the architecture's parameters.}.

The $m$ neighborhoods are used to structure the system inputs. 
As shown in Figure \ref{fig:feature_bank}, neighborhoods encode the network's input topology.
Since they make no assumption about an external space, or a geometry that relates these variables, neighborhoods can approximate many kinds of observation structures in principle.

A natural question is how the neighborhoods are specified. 
In general, without any prior structure given to the learning system, the neighborhoods could be chosen at random to provide nonlinear random projections~\cite{rahimi2008random}.  
The resulting neighborhoods do not reflect the structure of the observed data, but they can be useful in cases where a generic nonlinearity is sufficient.
In situations where there is some prior knowledge of the observation structure, the neighborhoods can be defined through an oracle side-channel.
Consider the situation where the observation vector is coming from irregularly-spaced sensors, and the data is driven by a spatial process. 
Each neighborhood can be defined by first selecting a sensor $\rho$, and then selecting the $k$ nearest sensors to $\rho$ in a distance-sorted order.

%% file: algorithm.tex
\section{Prediction Adapted Neighborhoods}
 We now turn to the question of how an RL system could learn to specify neighborhoods from statistics of its observation stream.
 Results from the predictive state literature show that a collection of predictions can be sufficient for representing state~\citep{littman2001,boots2011closing}.
 A separate line of prior work shows how informative spatial embeddings can be defined from statistics between different observation components~\citep{pierce1997map,tenenbaum2000global,roux2007learning,modayil2010discovering}.
Together with the idea of learning many GVF predictions in parallel~\cite{sutton2011horde,modayil2014multi,schaul2015universal}, we propose that useful neighborhoods could derive from statistics taken from auxiliary predictions of observation components.
In this section, we describe an algorithm that leverages this insight: computing neighborhoods from many GVF prediction objectives.
Since the selection matrices $\Mbf^i$ are adapted from the auxiliary predictions, we call these {\em prediction adapted neighborhoods}.

The computation of prediction adapted neighborhoods is outlined in Algorithm~\ref{alg:pan}, lines 6--14.
In essence, many GVFs are learned in parallel to form the matrices $\Mbf^i$.
Each GVF is the expected return of some signal $\bar{R}^i_{t+1}$, known as the \textit{cumulant}.
A cumulant's return is learned under the same policy and discount as the main value function \eqref{eq:value}, analogous to the return in Equation~\ref{eq:return},
 \begin{equation}
   \bar G^i_t \equiv \bar{R}^i_{t+1} + \gamma \bar{R}^i_{t+2} + \gamma^2 \bar{R}^i_{t+3} + \ldots.
 \end{equation}
The GVF cumulants are specified with a selector function, $c(i)$, that returns an index into the observation vector, such that $\bar{r}^i_{t+1} \equiv \obf_{t+1}[c(i)]$, for $i=1,\cdots,m$.
Each GVF objective has a separately learned solution, represented as a linear function of the observation inputs $\bar \xbf_t \equiv \obf_t$ and weights $\bar\wbf^i$: $\bar{\wbf}^{i\top}\bar{\xbf}_t\approx \mathbb{E}[\bar G^i_t|S_t=s]$.  
The weights are updated with TD$(\lambda)$, along with the associated eligibility traces $\bar\zbf^i$.
The weights are used to identify the relevant features for each GVF.
In particular, given the GVF weights $\bar\wbf^i$, the algorithm computes $\Mbf^i$ to select the $k\leq d$ observations with the largest absolute weights (Algorithm \ref{alg:pan} lines 12--14). 
The selection of the largest absolute weights can be expensive with generic implementations that use sorting as a subroutine, so for efficiency line 12 is optionally restricted to occur periodically. 

Given the neighborhood selection matrices, updating the main value function estimate (Algorithm \ref{alg:pan}, lines 16--18) relies on features $\xbf_t$, sparse connection imposed by neighborhood selection matrices, and feature parameters $\Abf$ (Algorithm~\ref{alg:stateupdate})\footnote{Prediction adapted neighborhoods should, in principle, be applicable in other settings than what is studied here.}.
As with the GVFs, the main value estimate is updated by TD($\lambda$), with the weights $\wbf$ and trace $\zbf$.

In contrast to prior work that employs auxiliary prediction to form better nonlinear features through regularization~\citep{jaderberg2016reinforcement}, the proposed algorithm \textit{indirectly} shapes nonlinear features by using information from the GVF weights to continually adapt the architecture.
Section \ref{sec:related_work} discusses connections to prior work on auxiliary learning in more detail. 

\begin{algorithm}[]
        \caption{Online Value Estimation with \\Prediction Adapted Neighborhoods}\label{alg:pan}
        \begin{algorithmic}[1]
        \STATE  Initialize parameters $\wbf$, $\zbf$,  $\Abf$, $\Mbf^{1:m}$,  $\bar \wbf^{1:m}$, $\bar \zbf^{1:m}$.
        \STATE Receive observation $\obf_1 $ from the environment.
         \STATE $\xbf_{1} \gets \text{ComputeFeatures}(\obf_{1},\Mbf^{1:m}, \Abf)$ 
        \FOR{$t=1,2,3,\cdots$}
        \STATE Receive $r_{t+1} , \obf_{t+1} $ from the environment.
        \tikzmk{A}
		\STATE $\bar \xbf_j \gets  \obf_{j} \textbf{ for } j \in \{t,t+1\} $
		\STATE \textbf{parallel for $i \in \{1,\cdots,m\}$}

	    \hspace{.95pc}{\color{comments} \# Update GVF weights with TD$(\lambda).$}
		\STATE \hspace{1pc}$\bar r^{i}_{t+1}\gets \obf_{t+1} [c(i)]$ 
	    \STATE \hspace{1pc}$\delta \gets \bar r^{i}_{t+1} + \gamma \bar{\wbf}^{i\top} \bar{\xbf}_{t+1} - \bar{\wbf}^{i \top} \bar{\xbf}_t$
        \STATE \hspace{1pc}$\bar{\zbf}^{i} \gets \gamma\lambda\bar{\zbf}^i + \bar{\xbf}_t$
        \STATE \hspace{1pc}$\bar{\wbf}^{i} \gets \bar{\wbf}^{i} +\bar\alpha \delta \bar{\zbf}^i$
		
		\hspace{0.95pc} {\color{comments} \# Construct top-$k$ selection matrix.}
		\STATE \hspace{1pc}$\ellbf\gets\text{Top}(k, \bar \wbf^{i}) \mbox{, where } { |\bar \wbf^{i}_{\ellbf_1}| \geq \! \cdots\! \geq | \bar \wbf^{i}_{\ellbf_k}| }$
        \STATE \hspace{1pc}\textbf{parallel for $j,l\in\{1,\ldots,k\}\times\{1,\ldots,d\}$}
		\STATE \hspace{2pc}$\Mbf_{j,l}^i\!\gets\!\mathbf{1}\{l=\ellbf_j\}$
        \STATE $\xbf_{t+1} \gets \text{ComputeFeatures}(\obf_{t+1},\Mbf^{1:m}, \Abf)$ 
        \tikzmk{B}
        \outlineboxit{black}
        
        {\color{comments} \# Update main prediction weights with TD$(\lambda).$}
        \STATE $\delta \gets r_{t+1} + \gamma \wbf^\top \xbf_{t+1} - \wbf^\top \xbf_t$
        \STATE $\zbf \gets \gamma\lambda\zbf + \xbf_t$
        \STATE $\wbf \gets \wbf +\alpha\delta \zbf$
        \ENDFOR
        \end{algorithmic}
\end{algorithm}

\begin{algorithm}[]
        \caption{ComputeFeatures}
        \begin{algorithmic}[1]
          \STATE \textit{input:}  $\obf, \Mbf^{1:m},  \Abf$
        	\STATE \textbf{parallel for $i \in \{1,\cdots,m\}$}
     	        \STATE \hspace{1pc}  $\ybf^i \gets  \fbf ( \Abf  \Mbf^{i} \obf )$
 	\STATE \textbf{return} $\text{concatenate}(\obf, \ybf^1, ... ,\ybf^m) $
        \end{algorithmic}\label{alg:stateupdate}
\end{algorithm}

%% file: experiments.tex
\section{Experiments and Results}\label{sec:experiments}
\subsection{How do GVF weights inform neighborhoods?}
\begin{figure}[t]
	\centering
	\includegraphics[width=\columnwidth]{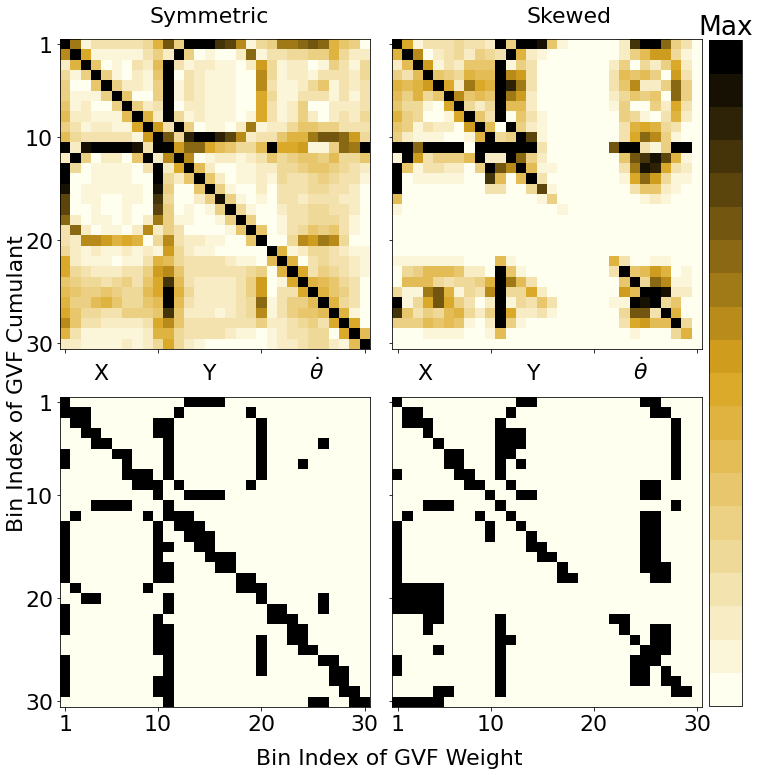}
	\caption{\textbf{Sparse structure from a simple pendulum:} GVF weights of discretized planar position and angular speed are shown for ten bins each (top row). The top-5 absolute values in each matrix row separate predictively-informative sparse structure from noise (bottom row). 
	The diagonal indicates most components are self-predictive.
	The symmetric policy induces symmetry in the weights of planar positions (see rings of high weight).
	These variables are most important for predicting each other.
	Symmetry breaks when the policy skews experience toward one side of the state space, causing angular speed to become more informative of the position.
	Weights are lower for high vertical positions, because these states are not often experienced under each policy.
 }
	\label{fig:pendulum}
\end{figure}
We begin with an example to illustrate the sparse structure GVF weights provide to neighborhoods.
Consider states from the simple pendulum in OpenAI Gym~\citep{brockman2016openai}, which include continuous planar position, $x,y$, and angular speed, $\dot{\theta}$.
Using ten bins to discretize the state dimensions, the observation vector has thirty binary-valued components, each reflecting the state's momentary occupancy of a bin.
From this, we define a GVF cumulant for each observation component.  The GVF prediction is approximated by a linear combination of the components (i.e. features).
We ask which components (bins) as features are important for predicting the other components as cumulants.

Similar to simple linear regression, prediction weights reflect the importance of each feature.
High magnitude weights suggest more importance than those with low magnitude. 
Unlike linear regression, however, GVFs predict a sum over a discounted future; so their high-weight features reflect an importance that extends over multiple time steps.

Figure \ref{fig:pendulum} plots a matrix of absolute weight values for all thirty GVFs.
Data is shown for a uniform random policy, that symmetrically samples experience around the pendulum's low-energy equilibrium, and a policy that skews the experience to one side of the equilibrium. 
In each case, the weights reveal a sparse structure which is identified with the top-5 absolute values, and which are used to define prediction adapted neighborhoods.
\begin{figure}[]
	\centering
	\includegraphics[width=.9\columnwidth]{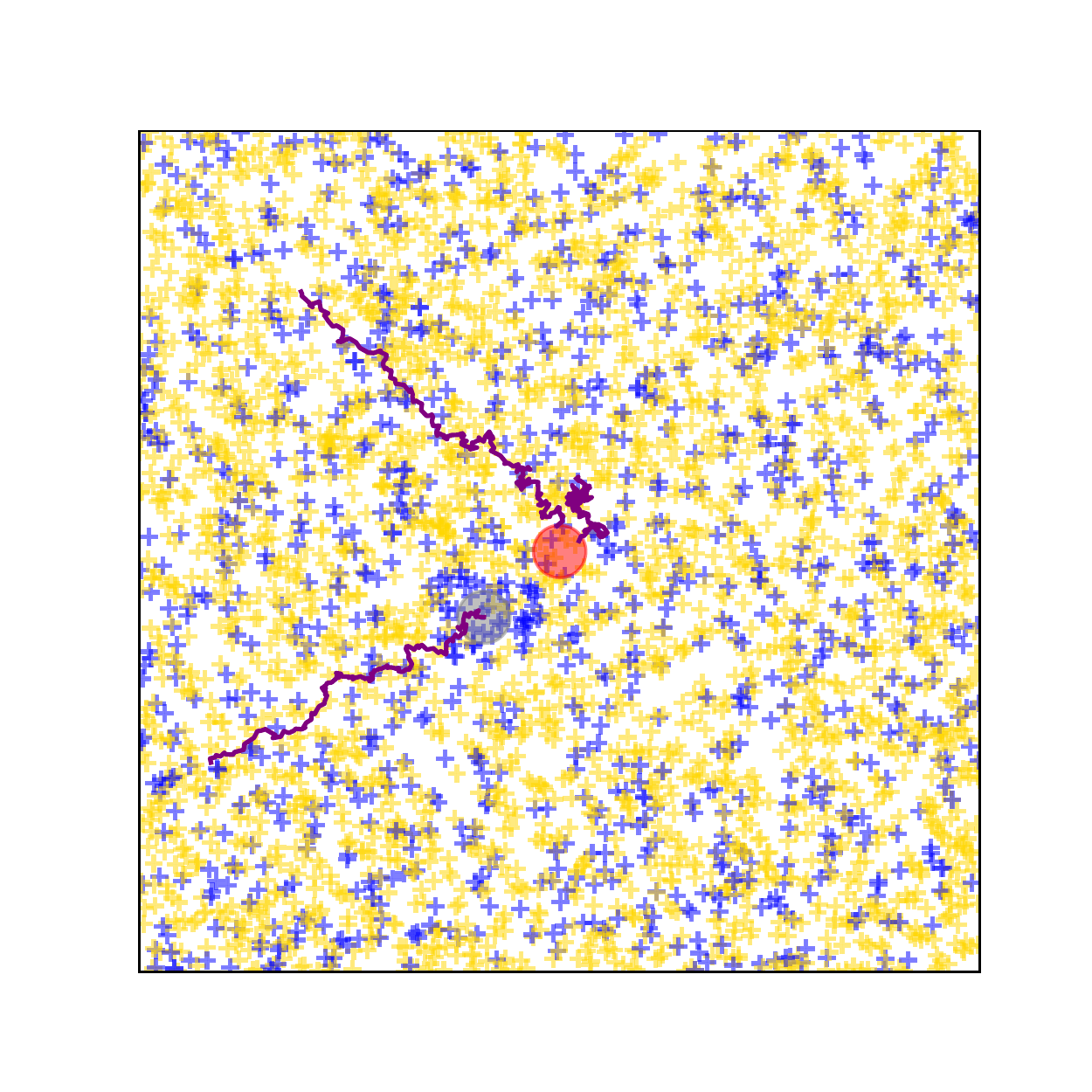}
	\caption{\textbf{The Frog's Eye domain:} An insect (gray circle) is detected by 4000 irregularly-distributed binary proximity sensors with considerable noise (blue: on, gold: off). Three insect trajectories are shown: two prior, one active. A reward of +1 is received when the insect enters the red region.}
	\label{fig:frogseye}
\end{figure}

\begin{figure*}
  \centering
	\begin{subfigure}[b]{0.3\textwidth}
		\centering
		\includegraphics[width=\textwidth]{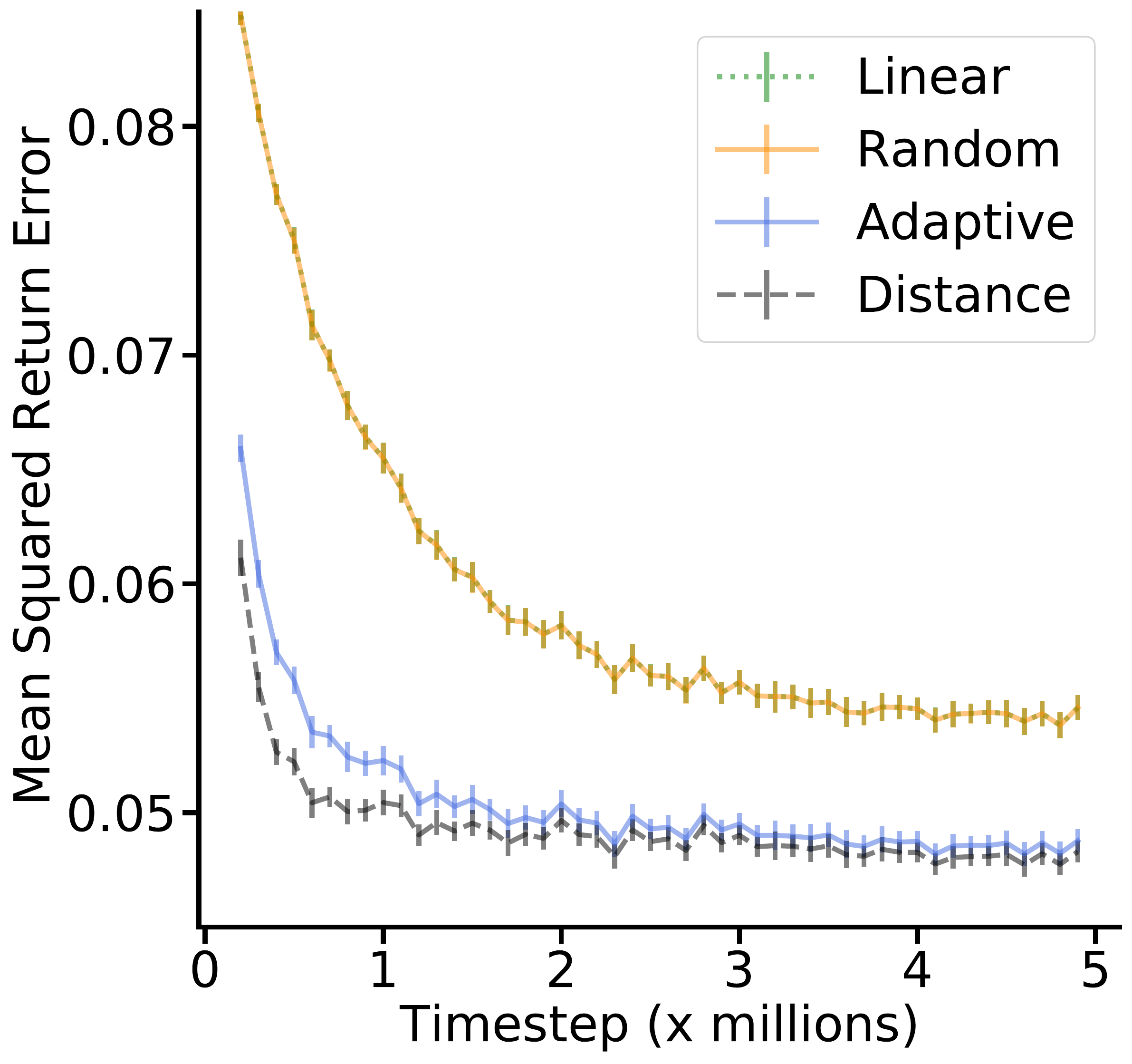}
		\caption{\footnotesize{Majority}}
	\end{subfigure}
        \hfill
	\begin{subfigure}[b]{0.3\textwidth}
		\centering
		\includegraphics[width=\textwidth]{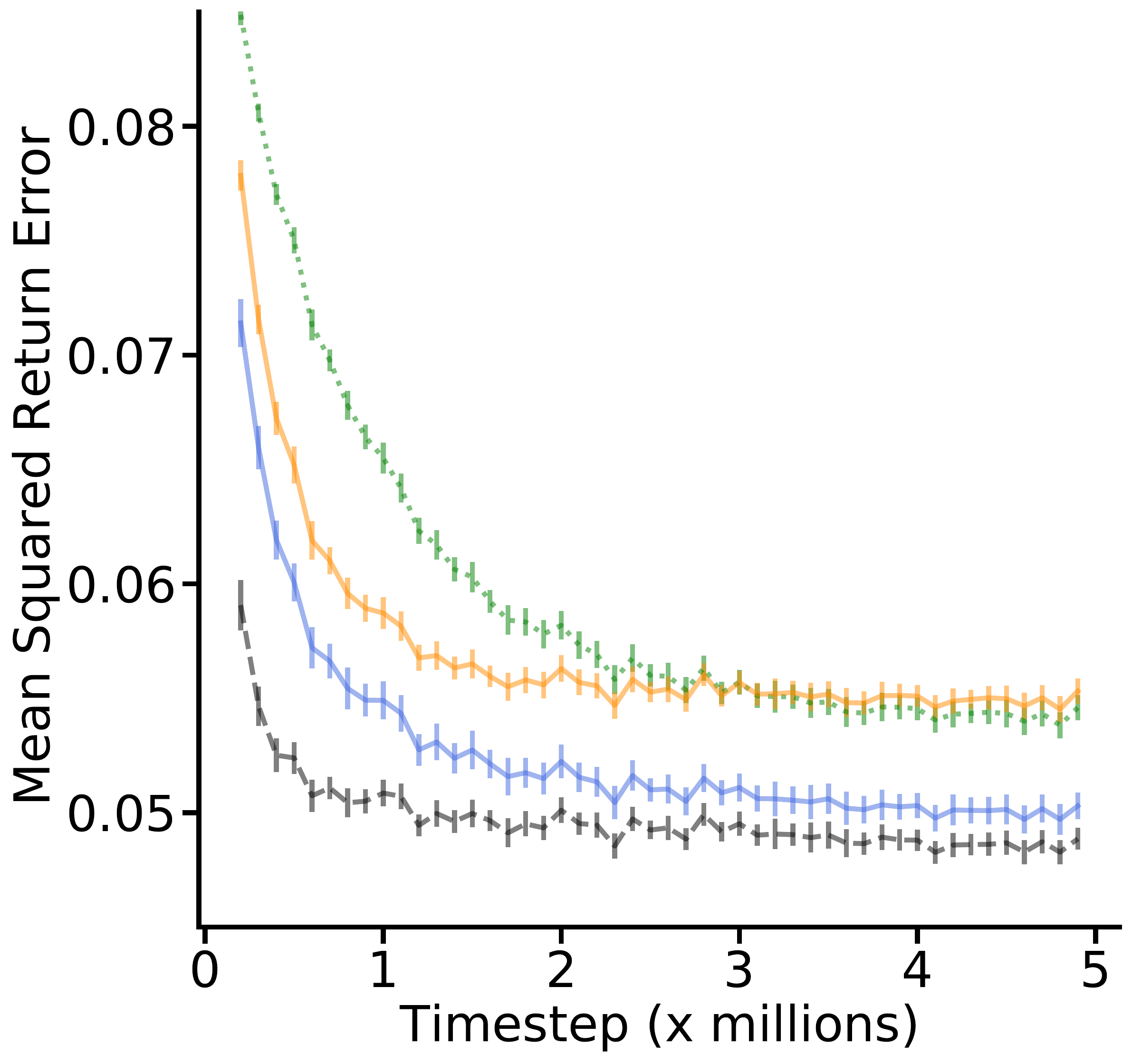}
		\caption{\footnotesize{LTU}}
	\end{subfigure}
        \hfill
	\begin{subfigure}[b]{0.3\textwidth}
		\centering
		\includegraphics[width=\textwidth]{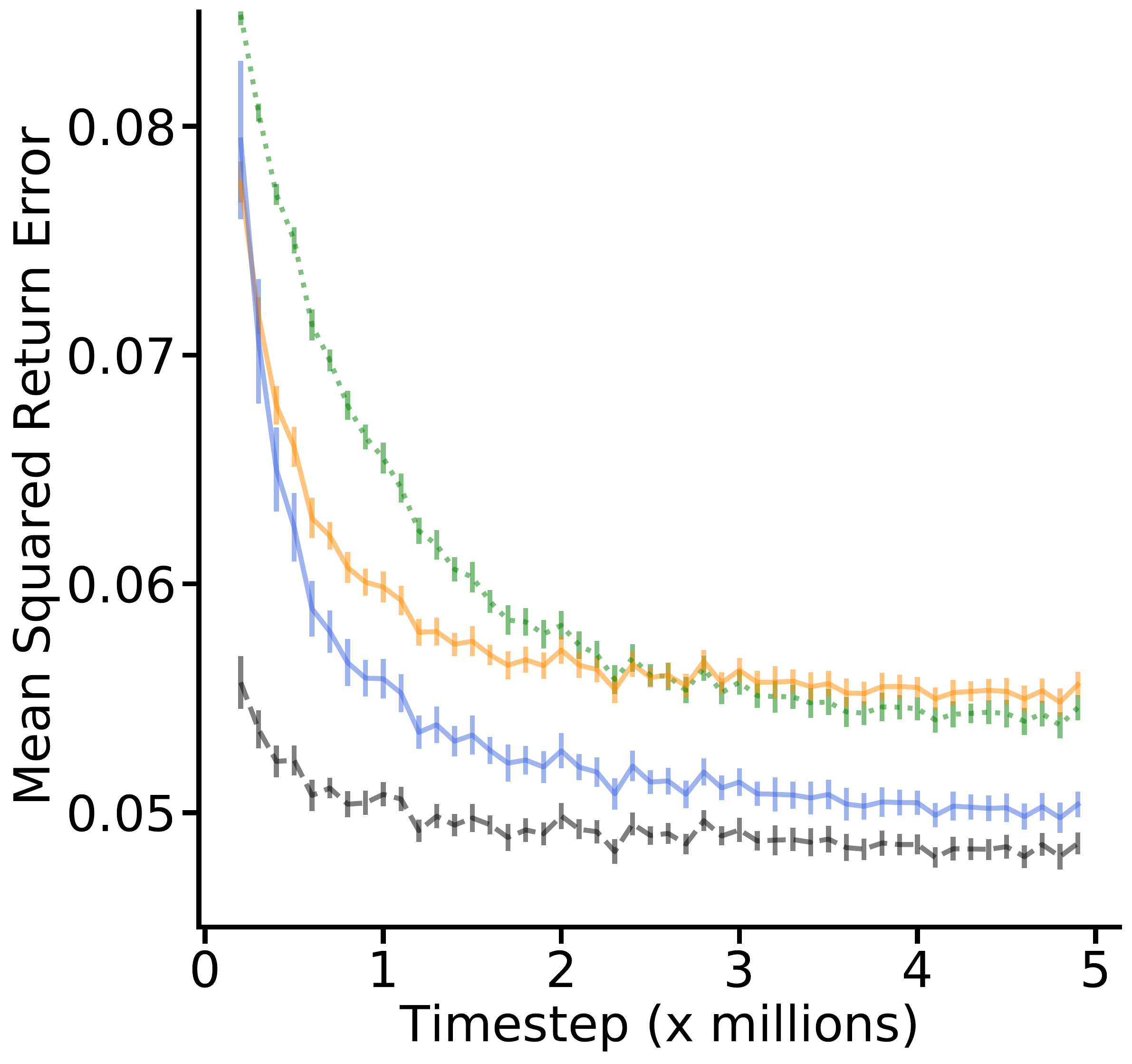}
		\caption{\footnotesize{ReLU}}
	\end{subfigure}
	\caption{\textbf{Prediction adapted neighborhoods are useful:} Training curves of prediction accuracy are shown for several approximation architectures configured with different neighborhoods and filter types. For each filter type, the Adaptive architecture (using prediction adapted neighborhoods) approaches the performance of the Distance architecture (using oracle information about the observation structure). Adaptive also performs significantly better than static architectures using random neighborhoods (Random) or none at all (Linear).
        }
	\label{fig:main-pred-performance}
\end{figure*}

\subsection{The Frog's Eye Domain}\label{sec:frogseye}
We now turn to a larger example, to study the full prediction problem.
We introduce an environment inspired by the functionality of a frog's eye, and the physical arrangement of its light receptors. 
Within a frog's eye, light receptors have a uniformly-irregular spatial distribution, with neither the concentrated fovea of a mammalian eye nor the regular grid of a silicon imaging chip.
A frog's eye is also known to induce a strong neural response when observing insect-like motion \cite{lettvin1959frog}.
In this environment, a simulated frog needs to anticipate the arrival of an insect without any knowledge of the underlying observation structure (i.e. how its light receptors relate to one another).

The simulated frog passively senses a two-dimensional space $\Pcal \equiv [-B,B] \times [-B,B]$ with 4000 binary proximity sensors (Figure \ref{fig:frogseye}). 
These sensors are uniformly-distributed at random across $\Pcal$ to mimic light receptors of a physical frog's eye. 
As the RL agent, the frog's objective is to learn a value function to predict the sum of future discounted reward, where reward is received when it captures an insect.

The full observation vector $\obf_t\in\{0,1\}^{4000}$ is given from a random ordering of all 4000 sensor readings. 
The observation's ordering is scrambled at the start of a trial and then held fixed. 
Observations are corrupted with noise\footnote{Observation noise is essential for studying nonlinear features, as a linear function of the observations would be highly informative in the absence of noise.} that changes sensor values to zero or one with a respective probability of $\epsilon/2$, where $\epsilon=0.5$.

The reward is one whenever the frog captures an insect, and it is zero otherwise. 
The positive reward comes after the insect enters a small circular region $\Rcal$ at the center of the observable space.  

The insect's position is given by a two-dimensional state $P_t\in \Pcal$ that changes by Brownian motion with an attraction toward the origin: $P_{t+1} = (1-\eta)\xi_{t+1}$, where $\xi_{t+1} \sim \Ncal(P_{t}, \sigma_P)$.
The attraction rate is set to $\eta=0.01$. The insect's initial position $P_1$ is assigned uniformly at random according to the distribution $ \Ucal(\Pcal\setminus\Rcal)$. 
When the insect drifts into the reward region, or leaves the space $\Pcal$, it will disappear and a new insect will spawn at a random location in $\Pcal\setminus\Rcal$. 
This process continues indefinitely.

\subsection{Experimental Methodology}\label{sec:methodology}
Our experiments compare the performance of different value function approximation architectures and focus specifically on the downstream effects of different neighborhood choices. 
Experiments define performance as a value function's prediction accuracy during online learning. 
Mathematically, this is the squared error between the prediction and the truncated empirical return, computed from Equation~\ref{eq:return} for every time $t\in \{1,\cdots,T\}$ with $T-t$ remaining rewards. 
The squared error is averaged over segments of $100000$ time steps, and error from the last segment is discarded to remove any measurable bias from truncation with $\gamma=0.99$. 
All experiments were implemented using JAX \citep{jax2018github}, and all parameters are tabulated in the Appendix.

Comparisons are made with four types of neighborhood: empty neighborhoods with $m=0$ (denoted as Linear), neighborhoods with $k$ randomly-selected observation components (Random), prediction adapted neighborhoods from fixed randomly-selected cumulants (Adaptive), and neighborhoods containing the $k$-nearest sensors to each cumulant using side-channel distance information (Distance).  
These selections allow us to assess how useful the prediction adapted neighborhoods are for forming nonlinear features relative to baselines that are always available (Linear and Random), and relative to a distance-based oracle \footnote{The Distance baseline does not represent the `best' learning system for our experiments; rather, it represents an alternative system that uses expert knowledge that is useful for its predictions.}.

Having selected a neighborhood, the architecture's performance will also depend on its nonlinear filter banks.
Three different filters are considered here (Table~\ref{tab:features}).
The first is a single binary majority filter tuned for the high noise level in this domain ($n=1$); it returns one when two thirds of the sensors in the neighborhood are active.  
The second is a linear threshold unit (LTU~\cite{sutton1993online}), with filter parameters drawn from a standard Gaussian, and an activation threshold of 4 applied within an indicator function. 
Using the same filters as the LTU, the third applies a ReLU function to the output, instead of a thresholded indicator. 

\begin{table}[]
\centering
  \begin{tabular}{l|l|l}
    Features & $f(z)$  & $\Abf$ shape and values  \\ \hline 
    $\text{Majority}$ &  $ \text{I}(z > \frac{2k}{3})$ & $(1\times k)$ filled with one\\
    \text{LTU} & $ \text{I}(z > 4)  $ & $(n\times k)$ from $\mathcal{N}(0,1)$ \\
   \text{ReLU} & ReLU$(z - 4)$ & $(n\times k)$ from $\mathcal{N}(0,1)$
\end{tabular}    
\caption{Nonlinear filter types defined on a neighborhood.}
\label{tab:features}
\end{table}

\textbf{Hyperparameter tuning:} Several hyperparameters were tuned for the experiments. 
Our goal was to find a regime with measurable differences between the Linear and Distance architectures; this indicated that some nonlinear features within the proposed architecture were useful for the prediction problem.
Using Majority features, a coarse search led to the neighborhood size $k=10$ used in all experiments. 
This setting strikes a balance in the signal provided by local sensor activity and the information from distal sensors, which added noise.
The TD update was configured with $\lambda=0.8$ and applied to both the main and auxiliary predictions. 
The number of filters for LTU and ReLU configurations ($n=100$) was determined by examining performance of the Distance architecture, as were the filter thresholds.
The GVF step-size ($\bar\alpha = 3\times 10^{-6}$) was found from an initial coarse search, and was held constant across configurations in the experiments.
Finally, a hyperparameter sweep for the main step-size ($\alpha$) was used to find the best setting for all methods, with more details in the Appendix.
Effects of varying the number of auxiliary predictions $m$ is detailed in a dedicated experiment.\footnote{Later experiments with the hyperparameters showed that higher choices of $\lambda$ improved the performance of all methods.  However, across multiple choices, the relative ordering between methods and the shape of the learning curves remained unchanged.}
 \begin{figure*}
	\centering
	\includegraphics[width=\textwidth]{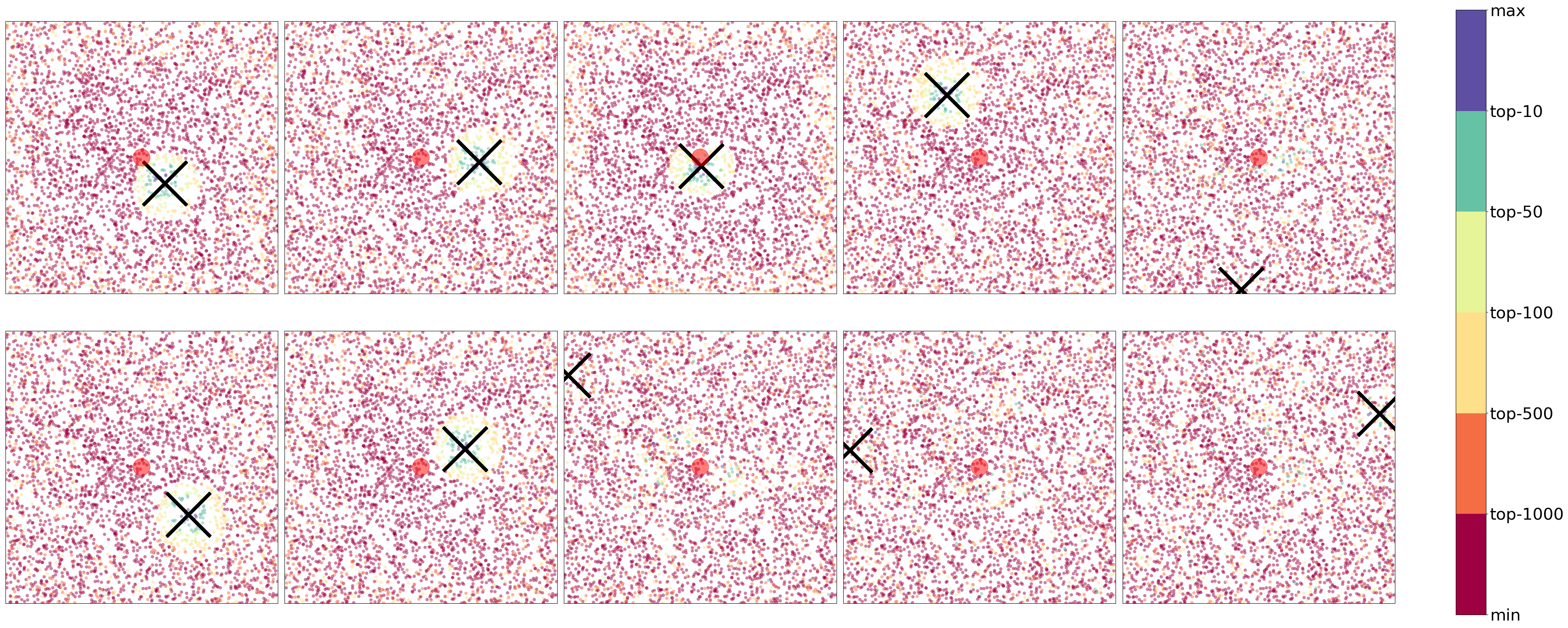}
	\caption{\textbf{Prediction adapted neighborhoods can encode locality in spatial domains:} The spatial distribution of GVF weights is shown for ten random cumulants ($\times$) after one million steps of training. Several top-$k$ neighborhoods are shown in different colors. The top 100 neighborhoods of several GVFs associate spatially-local sensors. Our experiments use the top ten to form nonlinear features.}
	\label{fig:aux-pred-weight-distribution}
\end{figure*}

\subsection{Are Prediction Adapted Neighborhoods useful?} 
Our first experiment addresses the question of whether the Adaptive architecture, with prediction adapted neighborhoods, provides measurable utility for the main prediction. 
The experiment follows the methodology described in Section \ref{sec:methodology}: comparing performance (i.e. prediction accuracy) of the considered approximation architectures.

Some algorithm parameters were held fixed during the experiment.
There were $m=4000$ neighborhoods in total: one for each observation component.
There were $n=100$ features computed from a sparse neighborhood of $k=10$ inputs. 
Architectures using LTU and ReLU filters produced $nm=400000$ complex features and the Majority filters produced $4000$ complex features.

We report the average and standard error confidence intervals from 30 trials.
The trials were run to 5 million steps, and were observed to complete in under two hours on a V100 GPU.

Performance data in Figure \ref{fig:main-pred-performance} allows us to draw several conclusions. 
Starting with the baselines, the data shows that Distance architectures, with their spatial neighborhoods, lead to the best asymptotic  performance. 
This result implies the domain is tuned appropriately for evaluating nonlinear functions of spatial features.
Furthermore, the steady state performance of using fixed random neighborhoods is statistically indistinguishable from using the Linear architecture.
Finally, the Adaptive architecture's resultant performance is similar to the Distance baseline; providing evidence that approximation architecture of a nonlinear value function can be adapted online and without significant performance loss.

\subsection{Does performance scale with more GVFs?} 
Our next experiment examines whether performance improves monotonically with the number of auxiliary predictions $m$.
Using the last 100000 measured steps of performance, this experiment compares the proposed Adaptive architecture with the Random and Distance baselines using Majority and LTU features for $m\in \{10,100,300,1000,4000\}$. These experiments report the average and standard error confidence intervals from 10 trials for each configuration.

\begin{figure}
	\centering
	\begin{subfigure}[b]{0.49\columnwidth}
		\centering
		\includegraphics[width=\textwidth]{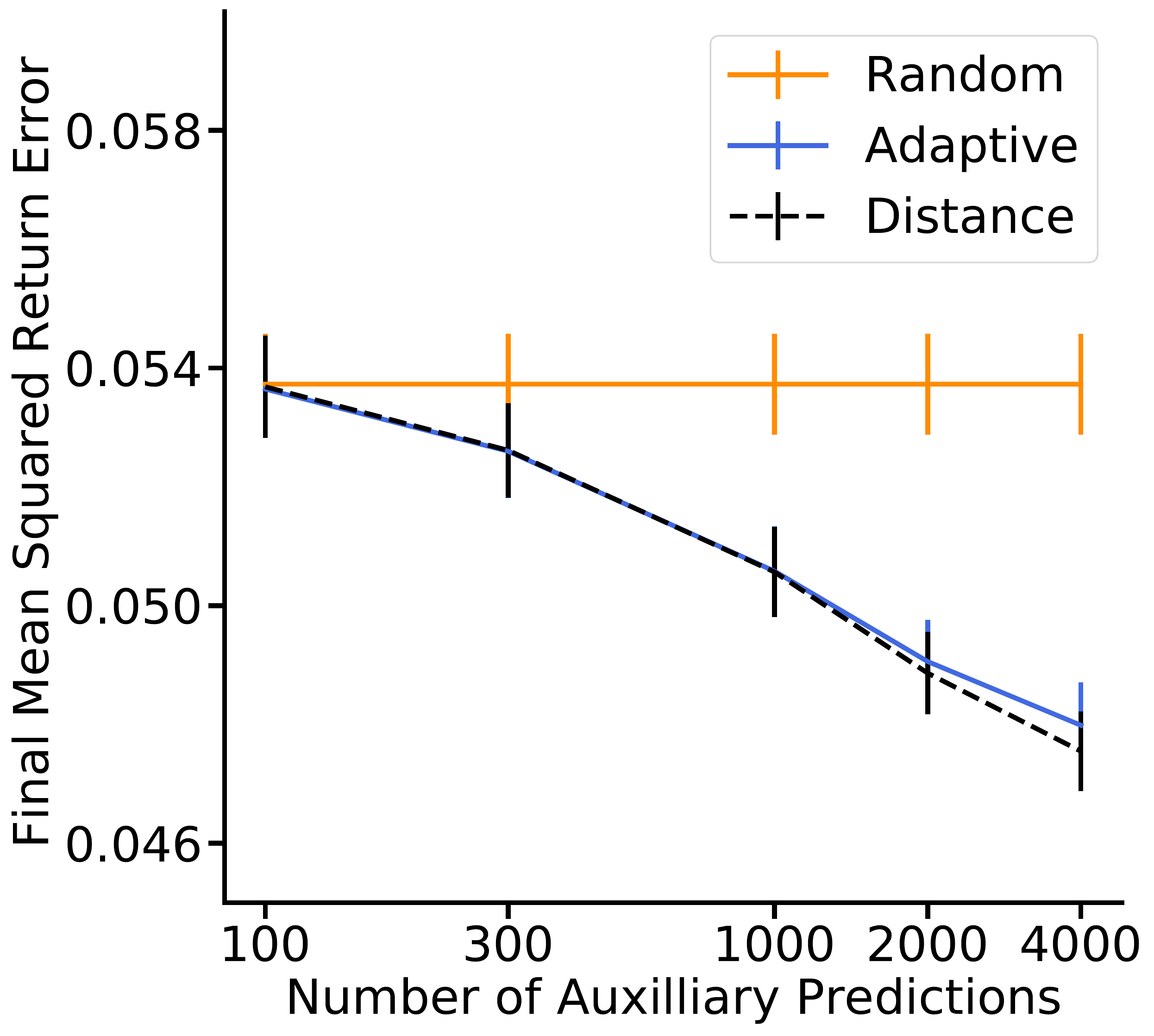}
		\caption{\footnotesize{Majority}}
	\end{subfigure}
	\begin{subfigure}[b]{0.49\columnwidth}
		\centering
		\includegraphics[width=\textwidth]{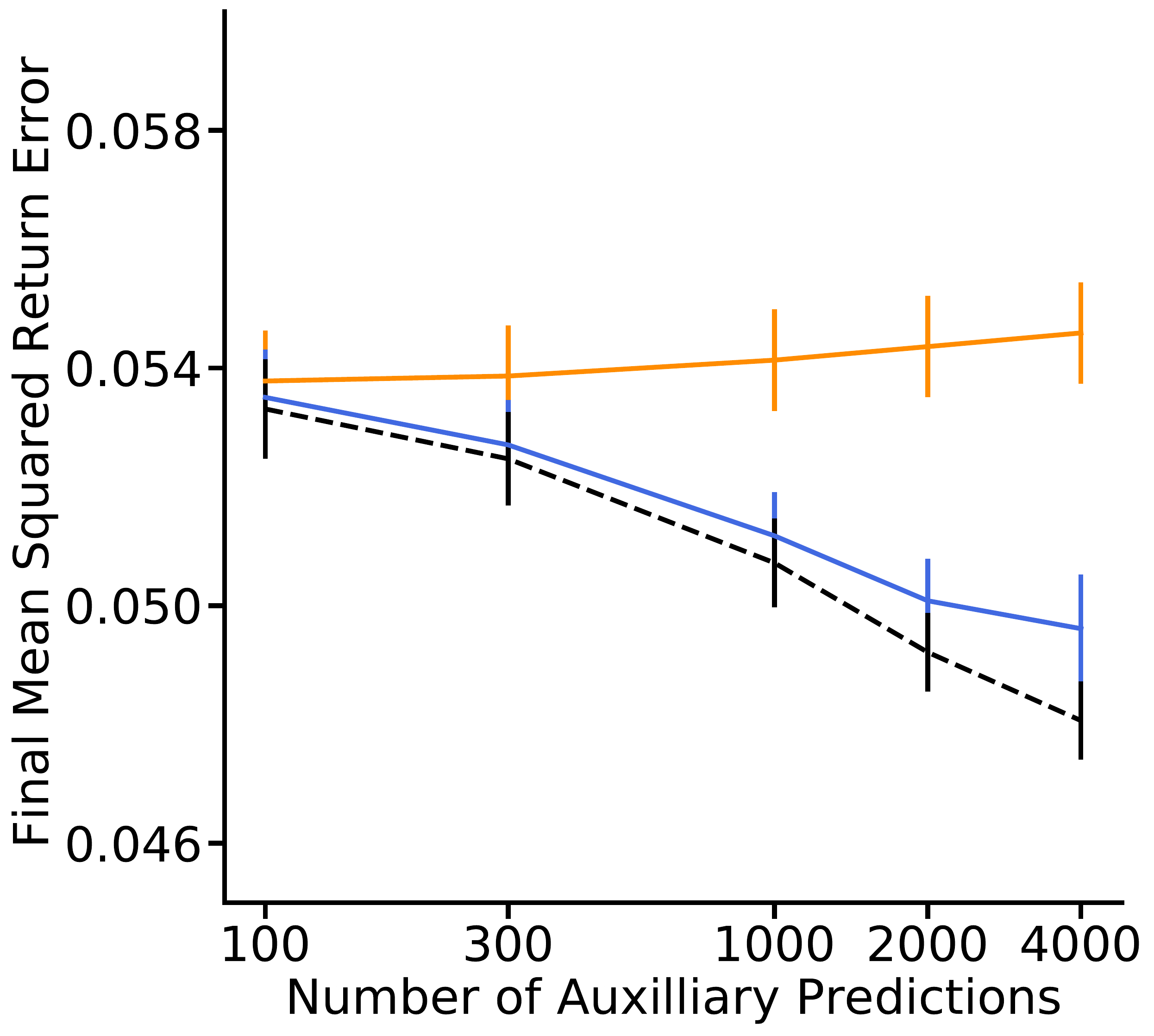}
		\caption{\footnotesize{LTU}}
	\end{subfigure}
	\caption{\textbf{Auxiliary Learning Effect}: Increasing the number of auxiliary predictions leads to performance improvements, as measured with final error.  The hyperparameters were chosen as shown in Table~\ref{tab:optim}. Standard errors are shown for 10 trials.
        }
	\label{fig:aux-pred-effect}
\end{figure}

The data in Figure \ref{fig:aux-pred-effect} indeed shows a performance improvement as $m$ increases. 
The best performance is with $m=4000$ predictions: one for each sensor in the simulated frog's eye. 
The data also reveals a lower limit, below which there is no significant benefit over the random baseline.  
The Adaptive architecture's performance scaling is comparable to that of the Distance baseline. 

These observations relate to two well-known results in RL.
Namely, the idea that states can be represented as a collection of predictions \cite{littman2001}.
And secondly, that a representation can improve with more auxiliary predictions \citep{jaderberg2016reinforcement}. 
As more GVFs are used to define additional prediction adapted neighborhoods, the approximations become better for representing the value function.

\subsection{Spatial structure of adapted neighborhoods}
\begin{figure*}
	\centering
	\includegraphics[width=\textwidth]{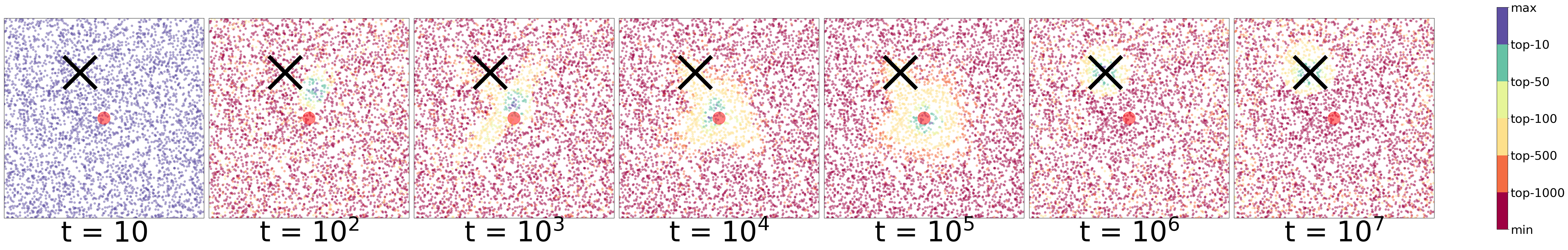}
	\caption{\textbf{Prediction adapted neighborhoods appear temporally stable:} We revisit the top-$k$ weight distribution from Fig. \ref{fig:aux-pred-weight-distribution} (fourth from top left) after ten-million steps of training. The neighborhood converges and remains centered around the cumulant's location ($\times$).}
	\label{fig:aux-pred-weight-distribution-series}
\end{figure*}

Given the comparable performance of spatially-fixed and prediction-adapted neighborhoods in Figure \ref{fig:main-pred-performance}, 
we examined the prediction adapted neighborhoods to see if they exhibited any spatial structure. 
Ten randomly-selected auxiliary predictions have their weights, after one million steps of learning, shown in Figure  \ref{fig:aux-pred-weight-distribution}.  The sensor location of the top-$k$ weights are shown by different color bands, for $k\in\{10,50,100,500,1000,4000\}$. 

Several properties can be observed in the images of Figure \ref{fig:aux-pred-weight-distribution}.
First, clustering around the cumulant sensor is observed in six of the ten weight sets.
Of those in the top 10, 50, and 100, there is similar spatial locality to what one would expect using nearest neighbors selection. 
Weak clustering around the bordering sensors is observed in the top 500 weights.
Note in cases when the insect is near the border, it will likely induce a new insect to spawn closer to the rewarding region. 
Uniform dispersion is seen in the remaining sets. 
No strong clustering is observed in four of the ten images, where the cumulant sensors are near the borders. 
We expect these predictions to train slower than others, because their cumulants will be mostly noise.

\subsection{Temporal stability of adapted neighborhoods}
A final inspection is performed with the auxiliary weights to examine whether their values are temporally stable over ten million time steps.
This inspection tracks one set of auxiliary weights at time steps that change at each order of magnitude: from $t=10,10^2,\cdots,10^7$.
The data is visualized in Figure \ref{fig:aux-pred-weight-distribution-series}.
The same cluster appears to stabilize some time after 100000 steps, suggesting that much of the learning in the auxiliary predictions is completed early.

%% file: related_work.tex
\section{Related Work}\label{sec:related_work}

\textbf{Auxiliary Learning:} 
The use of auxiliary learning has been used in RL to inform construction of the value function.
The Successor Representation (SR) \citep{dayan1993improving} is perhaps the earliest example, where predictions of the auxiliary cumulants $\bar{r}^{i}_t = \text{I}(s_t = i)$, for $i\in\Scal$, comprise the feature vector. 
While the SR's auxiliary predictions reflect expected future state occupancy, our auxiliary predictions reflect expected future observation activity. 
Auxiliary learning has also been applied to deep reinforcement learning systems, through additional loss terms, and it has shown to produce a regularizing effect when all the predictions share their features with the main value function \citep{jaderberg2016reinforcement, shelhamer2016loss}. 

\textbf{Computational Sparsity:}
Several offline and post-processing methods have been proposed for finding sparse network architectures. 
Approaches include constraining the network connectivity through connection pruning \citep{zhu2017prune}, regularization \citep{louizos2017learning,neyshabur2020towards}, and variational dropout \citep{molchanov2019pruning}. 
We find commonality with methods whose desired sparsity remains constant throughout training.
For instance, \citet{mocanu2018scalable} prunes and randomly introduces new connections to preserve the same sparsity. 
\citet{evci2019rigging} also prunes then introduces new connections based on gradient magnitudes. 
Our method does not dynamically change the connection count; it maintains a fixed neighborhood size, and it adapt the connections to select those which contribute the most to an auxiliary prediction. 
This technique shares similarity to the forward pass of Top-KAST \citep{jayakumar2020top}, which performs layer-wise top-$k$ masking based on feature weight magnitudes.

%% file: conclusion.tex
\section{Future Directions \& Summary}
In this work hidden layer parameters were held fixed, while the approximation architecture was adapted.
Although it was possible to learn useful predictions this way, it should still be beneficial to optimize a value function's internal parameters, as is the standard approach taken by deep RL algorithms.
In future research, it would be interesting to try back-propagation (or other methods) to optimize these parameters for further performance improvement.

Convolutional architectures can represent spatial patterns at multiple scales by composing the output from multiple layers. 
Our architecture examined only one hidden layer.
However, it may be useful to have more layers and adapt all the neighborhoods, or to study a recurrent architecture providing a similar functionality.
Such an architecture could facilitate information sharing and produce a richer set of features.

In this work, GVF questions were specified externally, from the observation.
It is natural to wonder whether additional benefit can be realized by internally generating GVF questions, perhaps by meta-learning~\cite{veeriah2019discovery}.

Another question is whether these ideas extend to the RL control setting.  
One challenge will be picking a good target policy for the GVFs when the behavior policy is changing. 
With that caveat, prediction adapted neighborhoods could be useful for approximating action values, and even within other data settings such as offline and batch RL.  

This paper addressed the problem of nonlinear prediction in RL and specifically considered architectures that could be updated in an incremental online fashion, and that were insensitive to observation structure. 
To this end, prediction adapted neighborhoods were proposed to change the network architecture for value function approximation.
The resulting architecture selects sparse subsets of the observations that have high predictive utility in separate parallel auxiliary predictions.
Empirical evidence was presented from two domains, showing that our adaptive architecture performs comparably well to a fixed architecture using a spatial bias, while remaining computationally tractable. 
Further evidence showed the presence of a new kind of auxiliary learning effect, where performance improved not as a result of regularization, but from adapting the approximation architecture. 
We believe this work could be useful for designing general RL systems that acquire knowledge from sensory inputs whose interdependencies are unknown.

%% file: acknowledgements.tex
\section*{Acknowledgements}
The authors would like to acknowledge the support of their many colleagues.
A special thanks goes to Tom Schaul, Zaheer Abbas, Brendan Englot, Paul Szenher, Dibya Ghosh, and Shruti Mishra for their comments on an early draft of this work; Parash Rahman for discussions on related work; Brian Tanner for his programming expertise; Michael Bowling, Patrick Pilarski, Rich Sutton, Adam White, and others at DeepMind for their comments and questions during the conceptual stages of this work.

%% file: supplement_abstract.tex
\newcommand{\myapptitle}{Adapting the Function Approximation Architecture \\ in Online Reinforcement Learning Appendix}
\icmltitlerunning{Appendix}
\onecolumn
\icmltitle{\myapptitle}

%% file: supplement_appendix.tex
\input{appendix}
\input{appendix_tables}

%% file: appendix.tex
\appendix
\section{Simple Pendulum Example Details}
The learning system was trained for 15,000 time steps using TD$(\lambda)$. 
The TD update was configured with $\gamma=0.9$, $\alpha=10^{-3}$, $\lambda=0.8$. 
Actions were chosen with two fixed policies; the symmetric policy, defined as the uniform distribution $\Ucal(-2,2)$, and the skewed policy $\Ucal(-1/2,3/2)$. 

The pendulum example's implementation is from OpenAI's Gym Environment.
The observations are given by discretization of horizontal and vertical positions, and angular speed. 
Each continuous variable is split into ten bins.
The positional coordinates vary between [-1,1], and the angular speed in [-8,8].

\section{The Frog's Eye Experimental Details}\label{sec:frogseye_appendix}
\subsection{Further domain definition}
The frog's eye is simulated in a square space $\Pcal\equiv [-B,B] \times [-B,B]$ centered at the origin with side length $2B=16$.
A particle with radius $r_P=0.5$ starts at a random location, $P_1$, drawn from the uniform distribution over the square $\Pcal$ and excluding the rewarding region $\Rcal$. 

The rewarding region is defined as the space enclosed by the disk of radius $r_R=0.5$ centered at the origin: $\Rcal \equiv \{P\in\Pcal \colon|| P || < r_R+r_P \}$. A reward of one is given exactly when the particle falls in this disk, $R_{t+1} = \text{I}(P_t \in \Rcal )$. The particle is reset at a random location, $P_{t+1} \sim \mathcal{U}(\Pcal\setminus\Rcal)$, if it previously fell inside the rewarding region or outside the square.

\subsection{Random seeds in the experiments}

In experiments with multiple trials, random seeds were shared across the tested components (for example, neighborhood selection mechanism in the FrogsEye domain), but were varied across each trial. This ensures a fair comparison between tested algorithmic components.  

The dynamics of the virtual fly and the locations of the sensors are set with a different random seed in each independent trial. 
The matrix $\Abf$, when drawn from a standard Gaussian, was selected independently on each trial. 

\subsection{Auxiliary cumulant selection}

Let $\Ccal$ be a set of $m\leq d$ indices, uniformly drawn without replacement from the initial set $\{1,\cdots,d\}$. Our experiments define the auxiliary cumulant selector function to be $c \colon \{1\cdots,m\}\rightarrow \Ccal$.

\subsection{Hyperparameter sweep}

\begin{figure*}
  \centering
        \begin{subfigure}[b]{0.3\textwidth}
                \centering
                \includegraphics[width=\textwidth]{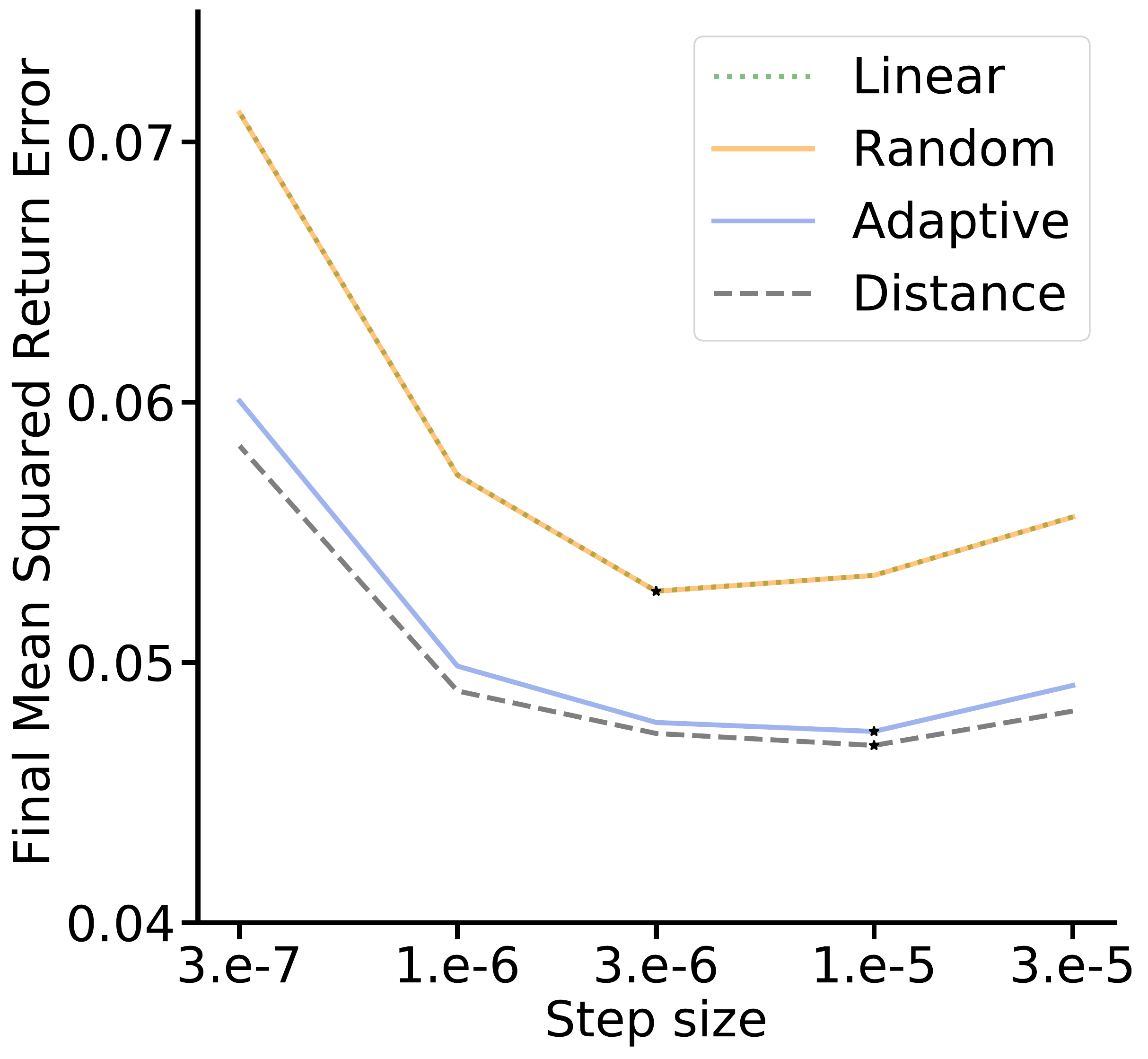}
                \caption{\footnotesize{Majority}}
        \end{subfigure}
        \hfill
        \begin{subfigure}[b]{0.3\textwidth}
                \centering
                \includegraphics[width=\textwidth]{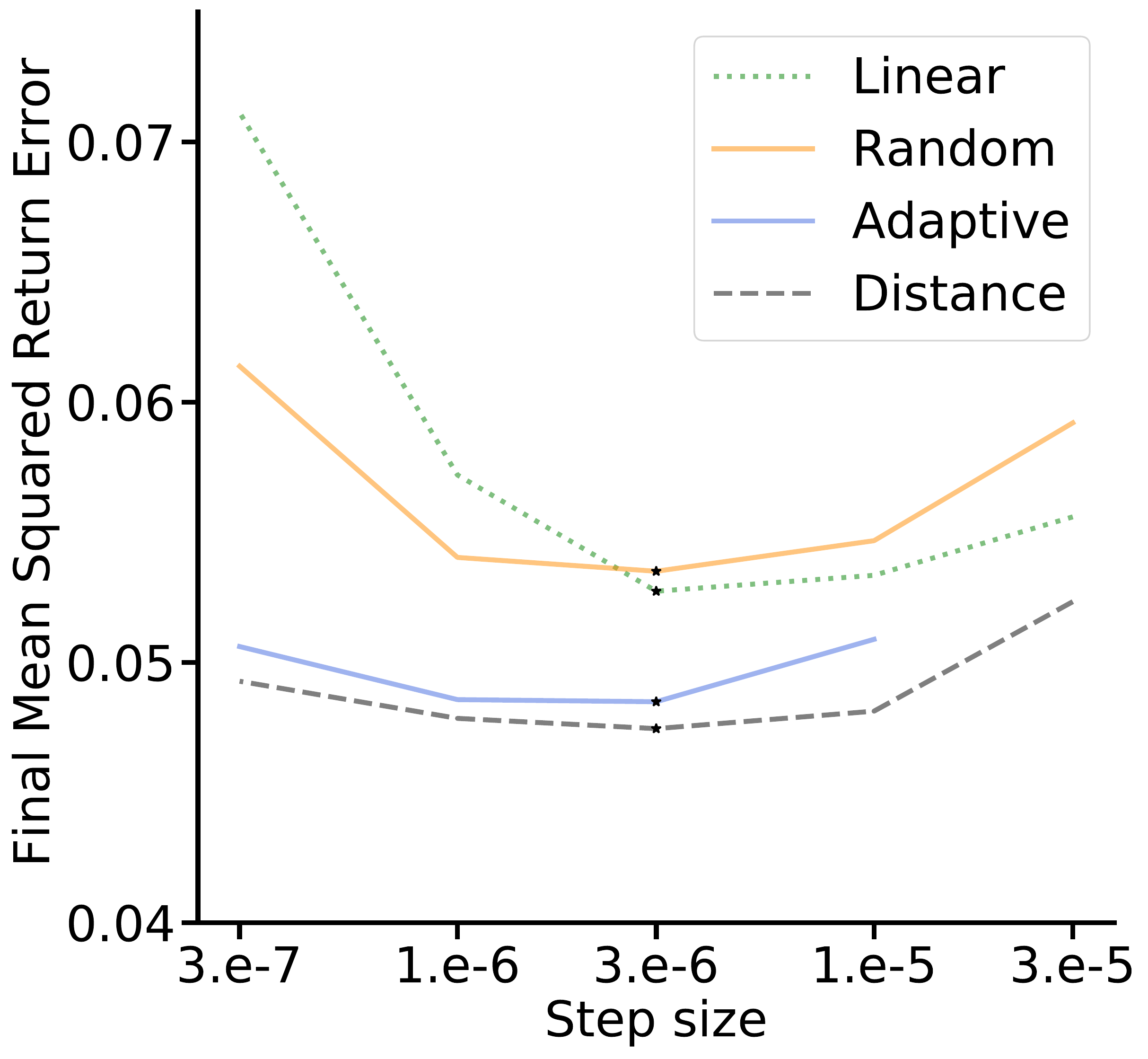}
                \caption{\footnotesize{LTU}}
        \end{subfigure}
        \hfill
        \begin{subfigure}[b]{0.3\textwidth}
                \centering
                \includegraphics[width=\textwidth]{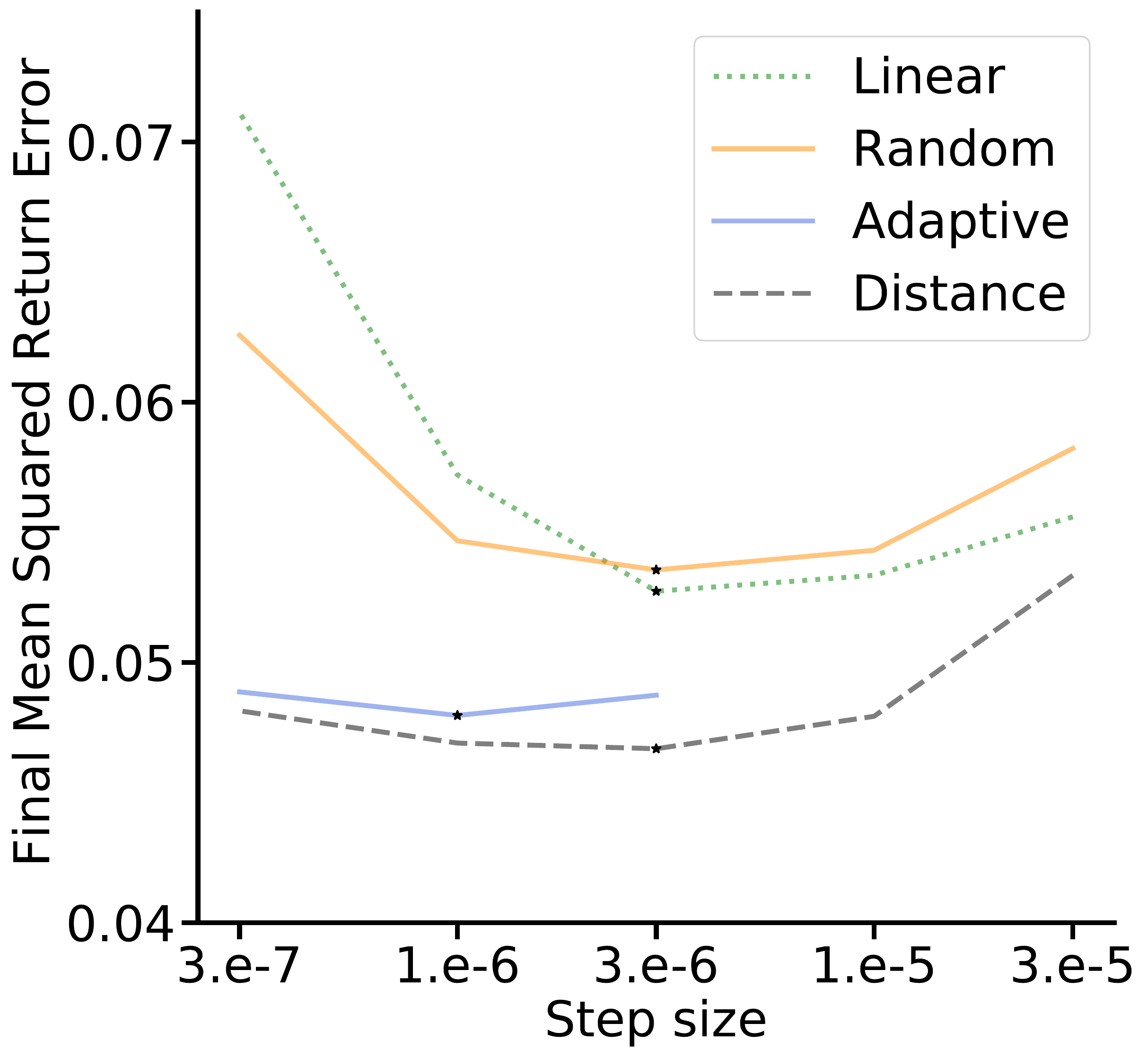}
                \caption{\footnotesize{ReLU}}
        \end{subfigure}
        \caption{\textbf{Step-size hyperparameter sweep:} For each of the three feature types, we performed a sweep over the step-size $\alpha$ for all the neighborhood types.  The final error was measured across three seeds for each configuration.  Note that the last $10^5$ time steps were reserved for computing returns and not evaluted, so the final error is the mean squared return error of the predictions over the previous $10^5$ steps (between $4.8$ and $4.9$ million time steps).  Averages that fell outside the plot were discarded.  We see the linear learner and the random neighborhoods have similar final errors, as do the distance-based neighborhoods and the adaptive neighborhoods.  The only configuration with substantial divergences was the combination of ReLU features with the adaptive neighborhoods. \label{fig:hyper}
        }
\end{figure*}

The step-sizes for the first experiment were selected by performing a coarse sweep over $\alpha$ over the values shown in Figure~\ref{fig:hyper}. 
\begin{table}[h]
\centering
\begin{tabular}{l|ccc}
Neighborhood & Majority & LTU &  ReLU \\ \hline
Linear (None) & $3\times 10^{-6}$ & $3\times 10^{-6}$& $3\times 10^{-6}$ \\
Random & $3\times 10^{-6}$ & $3\times 10^{-6}$ & $3\times 10^{-6}$  \\
Adaptive & $10^{-5}$ & $3\times 10^{-6}$ & $ 10^{-6}$ \\
Distance & $10^{-5}$ & $3\times 10^{-6}$ & $3\times 10^{-6}$
\end{tabular}
\caption{Best step-size ($\alpha$) for different nonlinear filters.}
\label{tab:optim}
\end{table}

\section{Complexity}
With many GVFs ($m=d$), computing the filters require approximately $kmd+kn=\Ocal(kd^2)$ floating points,
 and weights consume $md + mn + d =\Ocal(d^2)$. 
Assuming the matrix $\Abf$ is shared among the GVFs, this leads to a total memory footprint of $\Ocal(kd^2)$.
The learning system can require $\Ocal(kn)$ or $\Ocal(mn)$ when using many non-linear features, depending on the other factors.
The runtime of a neighborhood update contains a top-$k$ selection step which costs $\Ocal(d\log k)$  when implemented with a heap, though it is parallelized over the $m$ problems.
Updating the feature weights is linear in the number of features, and the full system maintains $md + mn + d$ total features.

%% file: appendix_tables.tex
\begin{table}
\centering
\begin{tabular}{l|l}
Parameter & Description  \\
\hline
${\bf o}\in \mathbb{R}^d$ & Observation vector\\
$ r \in \mathbb{R}$ & Reward\\
${\bf w}$ & Weights for main prediction \\
${\bf z}$ & Eligibility trace for main prediction \\
$\bar {\bf w}^{i}$ & Weights for $i$th auxiliary prediction \\
${\bf z}^i$ & Eligibility trace for $i$th auxiliary prediction \\
${\bf M}\in \mathbb{M}^{k\times d}$ & Neighborhood selection matrix \\
${\bf A} \in \mathbb{M}^{n\times k}$ & Fixed projection from a neighborhood \\
\end{tabular}
\caption{Notation for algorithms}
\label{tab:notation}
\end{table}

\begin{table}
\centering
  \begin{tabular}{l|l|l}
    Hyperparameter & Setting & Description \\ \hline
  $\alpha$ & $3\times 10^{-6}$ & Step size for the main prediction weights\\
  $\bar\alpha$ & $3 \times 10^{-6}$ & Step size for the auxiliary prediction weights\\
    $\lambda$ & $0.8$ & Eligibility trace parameter\\
    $d$ & 4000 & Number of sensors \\
$m$ & 4000 &  Number of neighborhoods/auxiliary predictions\\
$k$ & 10 & Dimension of each neighborhood \\
$n$ & 100 & Number of complex features in each neighborhood \\ 
    $h$ & 100 & Period of top weight selection\\
\end{tabular}    
\caption{Default Frog's Eye experiment hyperparameter settings.}
\label{tab:hyperparams}
\end{table}

\begin{table}
\centering
\begin{tabular}{c|c}
Parameter & Value \\
\hline
\# Sensors & 4000 \\
Sensor response radius & 0.6 \\
Space size & 16x16 \\
Discount factor $\gamma$ & $0.99$ \\
Noise rate  & 0.5\\
Attractive rate ($\eta$) & 0.01\\
Reward radius & 0.5\\
Particle radius & 0.5\\
Dynamics scale & 0.05
\end{tabular}
\caption{Frog's Eye environment dynamics parameters.}
\label{tab:frogseye-params}
\end{table}

\begin{table}
\centering
  \begin{tabular}{l|l}
Configuration &  Value \\ \hline
Number of Timesteps &   $5\cdot 10^6$\\ 
Wall clock time  bound &   2 hours\\ 
Number of seeds &  30 \\
\end{tabular}    
\caption{Other Frog's Eye experiment factors.}
\label{tab:experimental-facts}
\end{table}